\documentclass[lettersize,journal]{IEEEtran}
\usepackage{amsmath,amsfonts}
\usepackage{algorithmic}
\usepackage{algorithm}
\usepackage{array}
\usepackage[caption=false,font=normalsize,labelfont=sf,textfont=sf]{subfig}
\usepackage{textcomp}
\usepackage{stfloats}
\usepackage{url}
\usepackage{verbatim}
\usepackage{graphicx}
\usepackage{cite}
\usepackage{amssymb}
\usepackage{mathtools}
\usepackage{amsthm}
\usepackage[table]{xcolor}
\usepackage{tabularx}
\usepackage{multicol}
\usepackage{multirow}
\usepackage{booktabs}
\definecolor{grey}{RGB}{230,230,230}

\begin{document}

\title{DiffVC: A Non-autoregressive Framework Based on Diffusion Model for Video Captioning}

\author{Junbo Wang\IEEEauthorrefmark{1}, Liangyu Fu\IEEEauthorrefmark{1}, Yuke Li, Yining Zhu, Ya Jing, Xuecheng Wu, Jiangbin Zheng

\thanks{\IEEEauthorrefmark{1} Both authors contributed equally to this work.}

\thanks{Junbo Wang, Liangyu Fu, Yuke Li, and Jiangbin Zheng are with the School of Software, Northwestern Polytechnical University, Xi'an 710129, China (e-mail: jbwang@nwpu.edu.cn; lyfu@mail.nwpu.edu.cn; liyuke@nwpu.edu.cn; zhengjb@nwpu.edu.cn)}

\thanks{Yining Zhu is with the School of Computer Science, Northwestern Polytechnical University, Xi'an 710129, China (e-mail: yiningzhu@nwpu.edu.cn)}

\thanks{Ya Jing is with Beijing University Of Technology, Beijing 100124, China}

\thanks{Xuecheng Wu is with the School of Computer Science and Technology, Xi'an Jiaotong University, Xi'an 710049, China (e-mail: wuxc3@stu.xjtu.edu.cn)}

}

\markboth{Journal of \LaTeX\ Class Files,~Vol.~14, No.~8, August~2021}%
{Shell \MakeLowercase{\textit{et al.}}: A Sample Article Using IEEEtran.cls for IEEE Journals}


\maketitle

\begin{abstract}
Current video captioning methods usually use an encoder-decoder structure to generate text autoregressively. However, autoregressive methods have inherent limitations such as slow generation speed and large cumulative error. Furthermore, the few non-autoregressive counterparts suffer from deficiencies in generation quality due to the lack of sufficient multimodal interaction modeling. Therefore, we propose a non-autoregressive framework based on \textbf{Diff}usion model for \textbf{V}ideo \textbf{C}aptioning (DiffVC) to address these issues. Its parallel decoding can effectively solve the problems of generation speed and cumulative error. At the same time, our proposed discriminative conditional Diffusion Model can generate higher-quality textual descriptions. Specifically, we first encode the video into a visual representation. During training, Gaussian noise is added to the textual representation of the ground-truth caption. Then, a new textual representation is generated via the discriminative denoiser with the visual representation as a conditional constraint. Finally, we input the new textual representation into a non-autoregressive language model to generate captions. During inference, we directly sample noise from the Gaussian distribution for generation. Experiments on MSVD, MSR-VTT, and VATEX show that our method can outperform previous non-autoregressive methods and achieve comparable performance to autoregressive methods, e.g., it achieved a maximum improvement of 9.9 on the CIDEr and improvement of 2.6 on the B@4, while having faster generation speed. The source code will be available soon.
\end{abstract}

\begin{IEEEkeywords}
Video captioning, Diffusion model, Vision and language.
\end{IEEEkeywords}

\section{Introduction}
\IEEEPARstart{V}{ideo} captioning is a crucial task in vision and language. It aims to automatically generate accurate and coherent textual descriptions based on videos~\cite{venugopalan2015translating}, which requires the model to not only recognize objects, people, and scenes in the video, but also understand the temporal and spatial relationships, action sequences, and event developments between them, and ultimately convert visual information into a fluent sentence. Video captioning has broad application areas such as video understanding, video search, and recommendation systems. It is also an important benchmark task for evaluating the visual understanding and language generation capabilities of artificial intelligence systems.

\begin{figure}[t]
\begin{center}
   \includegraphics[width=0.99\linewidth]{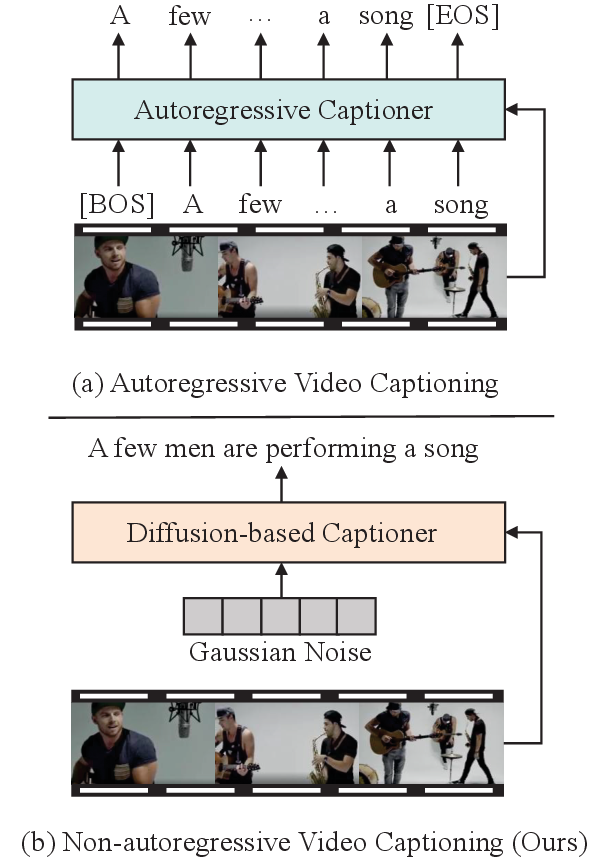}
\end{center}
   \caption{The sample comparison between (a) Previous diffusion-based video captioning approach and (b) Our proposed DiffVC.}
\label{fig: intro1}
\end{figure}

\begin{figure}[t]
\begin{center}
   \includegraphics[width=0.9\linewidth]{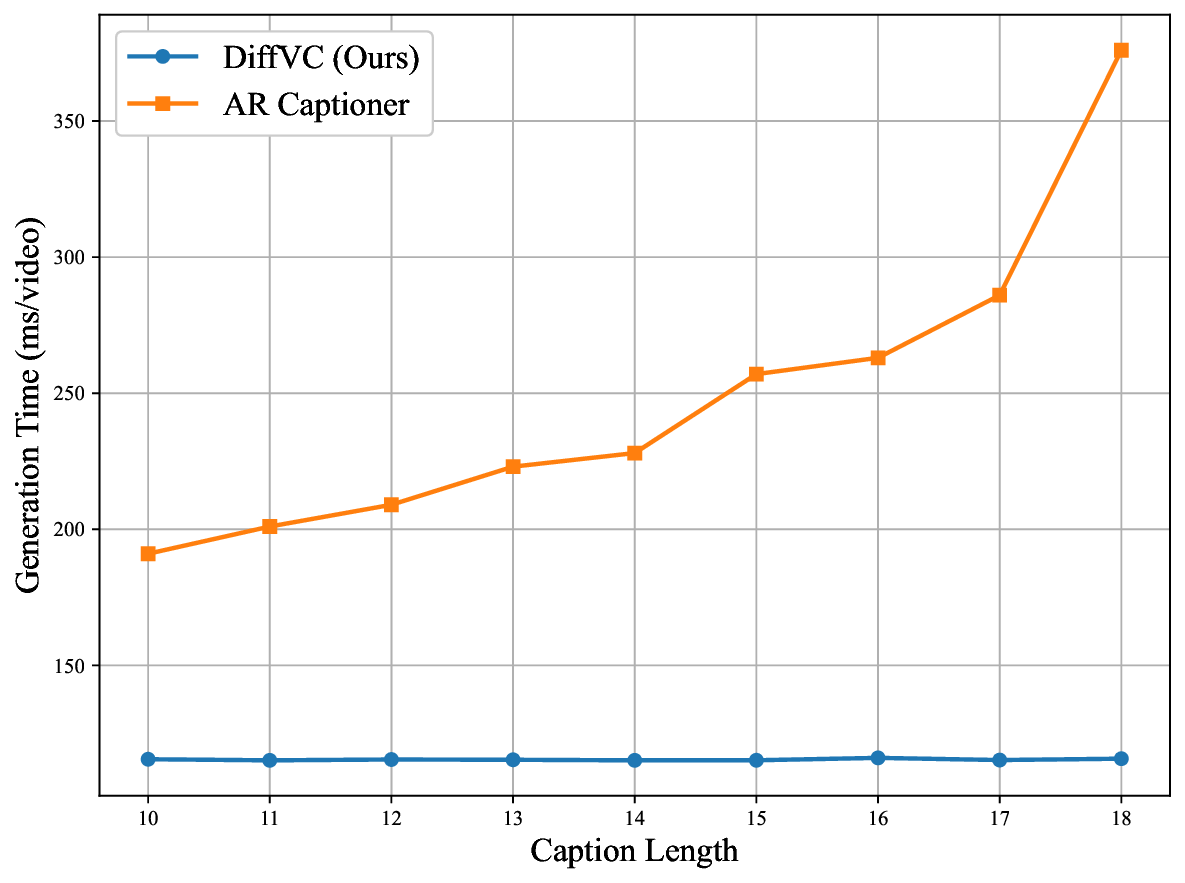}
\end{center}
   \caption{Comparison on generation speed between non-autoregressive and autoregressive video captioning. Autoregressive (AR) Captioner is KG-VCN \cite{yuan2025fully}.}
\label{fig: intro_speed}
\end{figure}

\begin{figure}[t]
\begin{center}
   \includegraphics[width=0.9\linewidth]{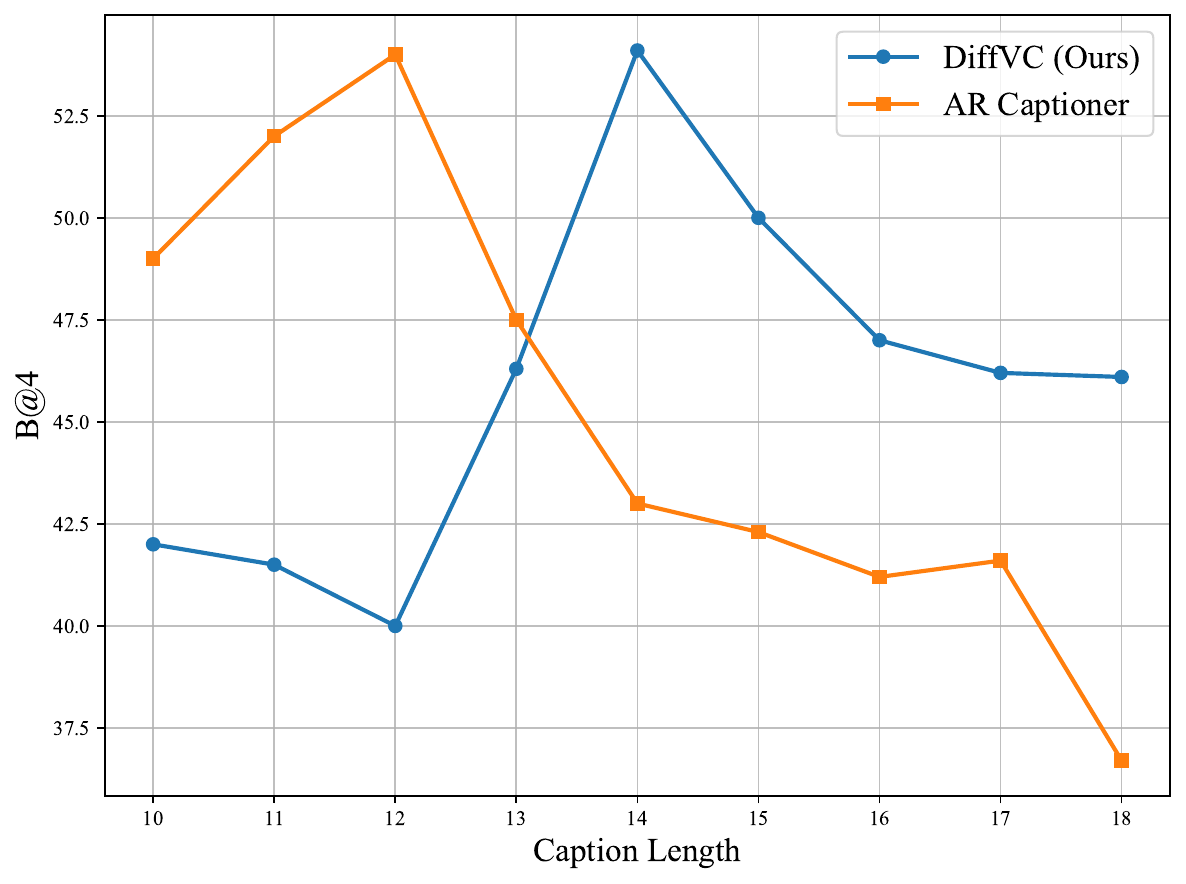}
\end{center}
   \caption{Comparison on generation quality between non-autoregressive and autoregressive video captioning, Autoregressive (AR) Captioner is KG-VCN \cite{yuan2025fully}.}
\label{fig: intro_b@4}
\end{figure}

Current video captioning methods generally follow the encoder-decoder paradigm, where the encoder encodes the input video into a visual representation, and the decoder autoregressively converts the visual representation into a textual description~\cite{chen2020videotrm, liu2018sibnet, wang2018reconstruction, wang2018m3}. Specifically, the decoder generates the next word based on the visual representation and the generated text content until the end symbol ([EOS]) appears, which is conducive to obtaining results with relatively accurate semantic logic. However, with the deepening of research, the inherent limitations of the autoregressive paradigm began to emerge: \textbf{(a) Slow generation speed:} As shown in Figure \ref{fig: intro_speed}, since the autoregressive paradigm generates words one by one, the generation time will increase significantly when generating long sentences, and the generation time is positively correlated with the sentence length. \textbf{(b) Cumulated error:} As shown in Figure \ref{fig: intro_b@4}, when the autoregressive paradigm generates each word, it uses the visual representation and the generated text content as a reference. If there are serious errors in the generated content, the subsequent generated words will be far from expectations, resulting in the loss of generation quality. This problem is particularly serious when generating long sentences.

As shown in Figure 1(b), non-autoregressive video captioning generates each word in parallel. For sentences whose actual length is less than the maximum generated length, the redundant tokens are replaced by masks. Non-autoregressive methods can effectively address these inherent limitations of autoregressive methods. However, previous non-autoregressive methods have disadvantages in generation quality. Due to the lack of modeling and discriminating correlation patterns between and within vision and language, the generated textual descriptions are prone to semantic problems, such as missing words.
To alleviate the above issues, we propose a diffusion-based framework for non-autoregressive video captioning, named DiffVC. Specifically, we first encode the input video into a visual representation using a spatiotemporal encoder. During training, we encode the ground-truth caption into a textual representation using a text encoder, and then gradually add Gaussian noise to the textual representation. Next, we propose a discriminative denoiser to gradually generate new textual representations from Gaussian noise using the visual representation as a conditional constraint. Finally, we decode the new textual representation using a language model to generate textual descriptions. During inference, since there is no ground-truth text involved, we directly sample a noise input from the Gaussian distribution into the denoiser, and then generate the corresponding textual description from the noise based on the visual representation.

In summary, the contributions of this paper are as follows: 

\begin{itemize}
    \item We propose a diffusion-based framework for non-autoregressive video captioning, addressing the inherent limitations of previous autoregressive counterparts such as slow generation speed, large accumulated error, and low diversity.
    \item We propose a discriminative denoiser to specifically model the inter-modal and intra-modal patterns between vision and language, so as to improve generation quality and address the semantic deficiencies of non-autoregressive methods.
    \item Extensive experiments on MSVD, MSR-VTT, and VATEX datasets demonstrate that DiffVC can achieve state-of-the-art non-autoregressive video captioning capabilities while being comparable to autoregressive counterparts.
\end{itemize}

\section{Related Work}
\label{sec:rel}

\subsection{Autoregressive Video Captioning}

Autoregressive video captioning generate sentences word-by-word \cite{venugopalan2015translating, liu2018sibnet}, the single word is conditioned on the previously generated words and visual content, the objective is to maximize the joint probability of the target words. \cite{aafaq2019spatio} presents a visual feature encoding technique to generate semantically rich captions. \cite{chen2019motion} presents a novel video captioning framework to learn spatial attention on video frames under the guidance of motion information for caption generation. \cite{chen2023retrieval} presents a Retrieval Augmentation Mechanism (RAM) that enables the explicit reference to existing video-sentence pairs within any encoder-decoder captioning model. \cite{li2022long} present the long short-term relation transformer to resolve issues such as redundant connections, over-smoothing, and ambiguity in relationships within video content. Additionally, \cite{deng2021syntax} present the syntax-guided hierarchical attention network to better combine visual and contextual features in captioning. However, this sequential word-by-word generation method has inherent limitations such as slow generation speed and large cumulative error.

\subsection{Non-autoregressive Video Captioning}

Non-autoregressive video captioning decodes all target words simultaneously, effectively overcoming the speed limitations associated with autoregressive counterparts. \cite{yang2021non} first proposed a non-autoregressive decoding based model with a coarse-to-fine captioning procedure. \cite{chen2024action} proposed the Action-aware Language Skeleton Optimization Network (ALSO-Net) tackles the challenge of
extracting action information across frames, improving understanding of complex context-dependent video
actions and reducing sentence inconsistencies. There are few studies on non-autoregressive video captioning, and the issue of generation quality needs to be addressed.

\subsection{Diffusion Model}

Early generative models are mainly implemented by generative adversarial networks (GAN) \cite{goodfellow2020generative} and variational autoencoders (VAE) \cite{kingma2013auto}, but these methods have limitations such as difficulty in training and mode collapse. The diffusion model effectively solves these problems, and its generation quality is also significantly improved, so it has become the current mainstream generative model. In image generation, \cite{ho2020denoising} used a diffusion probabilistic model to obtain high-quality images. \cite{rombach2022high} applied diffusion models in the latent space of a pre-trained autoencoder. Training diffusion models on this representation enables a trade-off between training cost and generation quality. \cite{podell2023sdxl} proposed SDXL, which expanded the scale of the model based on the previous stable diffusion models and greatly improved the generation quality. Based on SDXL, \cite{sauer2025adversarial} proposed Adversarial Diffusion Distillation (ADD) to reduce the inference time steps to 1-4 while maintaining the same image quality.

Based on the above research, diffusion models have achieved excellent performance in processing generation tasks of continuous data such as images and audio. However, generation tasks that process discrete data such as text, e.g. image/video captioning, are still challenging for diffusion models.

\section{Proposed Method}

Figure \ref{fig: overall}(a) shows the overall architecture of DiffVC. It is a diffusion-based non-autoregressive video-to-text generation framework. During training, the input contains two modalities: video and text. We use a visual encoder and a text encoder to encode the video and text into visual representation and textual representation, respectively. Next, we gradually add Gaussian noise into the textual representation, and propose a denoiser to generate a new textual representation from the noisy textual representation. Finally, we input the new textual representation into a non-autoregressive language model based on the Transformer to generate the caption. During inference, we use the visual encoder to encode the input video into a visual representation, and then sample noise from a Gaussian distribution. The noise and visual representation are input into the discriminative denoiser, it generates a new textual representation from the noise based on the visual representation. Finally, the language model generates the caption based on the new textual representation. The following is a detailed description of DiffVC: 

\begin{figure*}[ht]
\begin{center}
   \includegraphics[width=\linewidth]{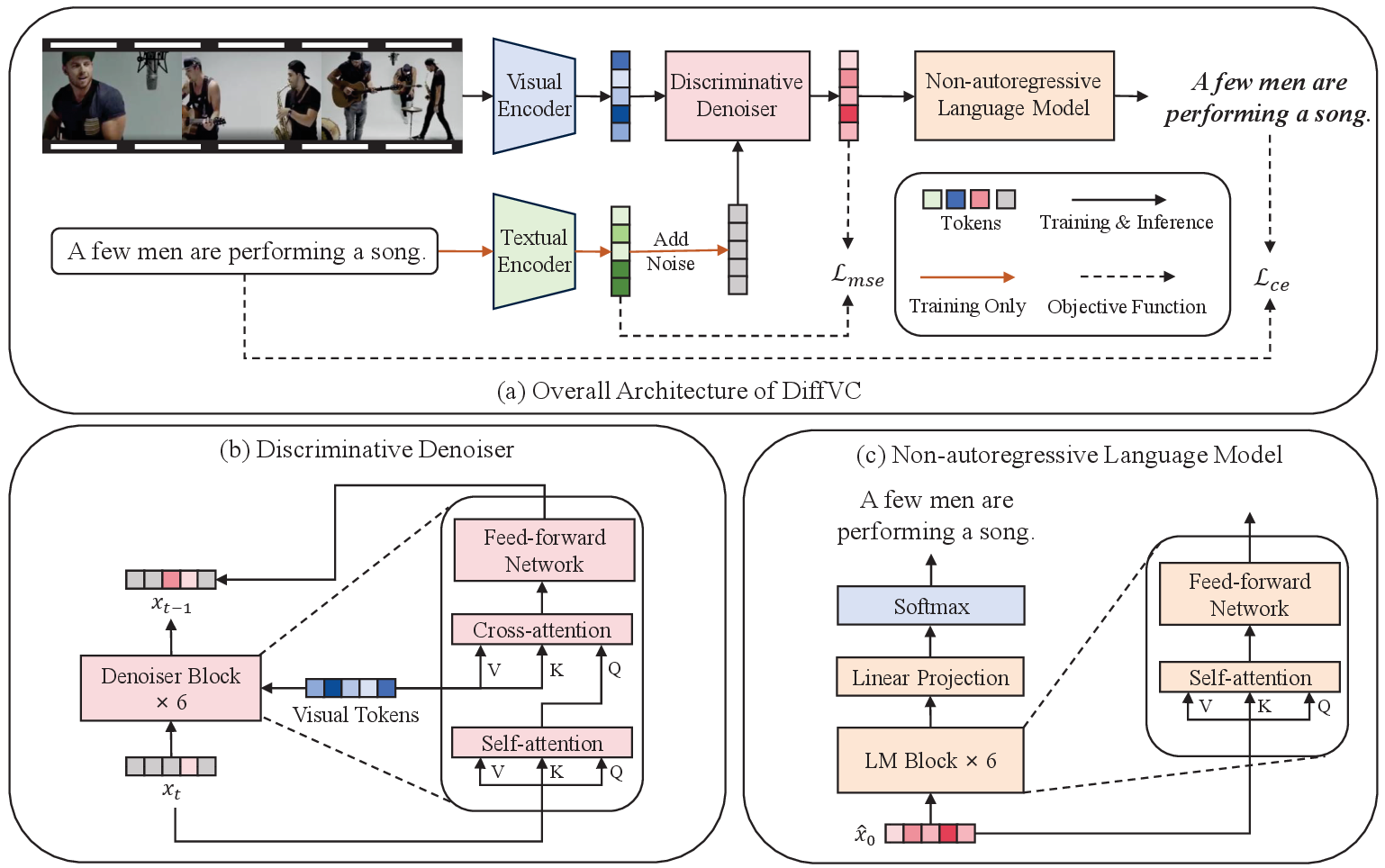}
\end{center}
   \caption{The overall architecture and key modules of our proposed DiffVC. (a) is the overall architecture of DiffVC, the string in italics denotes the generated caption of a video. (b) is the architecture of the discriminative denoiser. (c) is the architecture of the non-autoregressive language model.}
\label{fig: overall}
\end{figure*}

\subsection{Diffusion for Textual Representations}

Given the video $ i \in R^{N \times H \times W \times C}$, it is encoded into the visual representation $v$ by the pre-trained visual encoder from RSFD \cite{zhong2023refined}. Given the text (ground-truth captions) $ c \in R^{L}$, it is encoded into the textual representation $x_0$ by the pre-trained text encoder. Next, we gradually add Gaussian noise to $x_0$, and obtain a series of noisy textual representations $X=(x_1, x_2, …, x_T)$, where $x_t \in R^{N_v \times d_v}$,  $T\to \infty$ is a hyperparameter denoting the timesteps, $N_v$ is the number of tokens, and $d_v$ is the dimension of tokens; the forward diffusion process can be expressed as:

\begin{equation}
\begin{aligned}
x_t= & \sqrt{\alpha_t} x_{t-1}+\sqrt{1-\alpha_t} \epsilon_{t-1}, \\
\end{aligned}
\end{equation}
where $\alpha_t = 1 - \beta_t$, $\bar{\alpha}_t = \prod_{i = 1}^t \alpha_i$ and $\beta_t \in (0, 1)$  is a variance schedule. The probability density can be expressed as:

\begin{equation}
    q(x_t|x_{t-1}) = \mathcal{N}(x_t; \sqrt{1-\beta_t} x_{t-1}, \beta_t\mathrm{\textbf{I}}),
\end{equation}

\begin{equation}
    q(x_t|x_0) = \mathcal{N}(x_t; \sqrt{\bar{\alpha}_t} x_0, \sqrt{1-\bar{\alpha}_t}\mathrm{\textbf{I}}),
\end{equation}
where $\mathcal{N}(\cdot)$ is the Gaussian distribution and $\mathrm{\textbf{I}}$ is the identity matrix.

\textbf{Discriminative denoising.} For a series of noisy textual representations~$X$, we remove the noise that exists in them. In the backward denoising phase, we propose the discriminative denoiser $f_{\theta}$, its architecture is shown in Figure \ref{fig: overall}(b). 

In the original concatenation method of ‘[CLS] + textual representation’, the conditional constraints ([CLS]) and the text need to learn complex association patterns together in the self-attention, which can easily dilute the conditional information by the text’s own attention due to weight sharing and global interaction. The text can only passively refer to [CLS] in the global self-attention, and the adaptation rights and responsibilities of the features are not clear enough. 

To optimize the role of visual conditional constraints in the textual representation generation process and ultimately generate high-quality textual descriptions, we designed a discriminative denoiser block in the denoiser. By independently dividing the interaction path of the condition (Key/Value) and the text (Query) across attention, the model can explicitly distinguish the functions and semantics of the two, avoiding the problem of the condition being submerged in the self-attention layer. At the same time, it allows the query at each text position to calculate the correlation with all the features (Key) of the condition separately. For example, the verb in the text may focus on the action semantics in the condition; the noun may focus on the object description in the condition. 

Specifically, the discriminative denoiser accepts two inputs: a noisy textual representation and a visual condition constraint. The noisy textual representation is used as the query vector in the attention after a linear projection, and the visual condition constraint vector is used as the key vector and value vector after two linear projection, respectively. After that, the QKV vector is passed through the cascade denoiser blocks to generate a new textual representation $\hat{x}_0$ for the language model. The probability density of the backward denoising process can be expressed as:

\begin{equation}
p_\theta\left(x_{t-1} \mid x_t\right)=\mathcal{N}\left(x_{t-1} ; \mu_\theta\left(x_t, t\right), \Sigma_\theta\left(x_t, t\right)\right),
\end{equation}
where $\mu_\theta\left(x_t, t\right)$ and $\Sigma_\theta\left(x_t, t\right)$ are parameterized by BERT. In general, the whole backward denosing process can be summarized as:

\begin{equation}
    \hat{x}_0 = f_\theta \left( x_t, v, t \right),
\end{equation}
where the $\hat{x}_0$ is the generated textual representation, $\theta$ is the parameter of the discriminative denoiser, $v$ is the visual conditional representation encoded by the visual encoder, and $t$ is the timestep.

\subsection{Non-autoregressive Language Model}
\label{sec: lm}

Finally, we proposed a non-autoregressive language model to generate captions based on textual representation, the architecture of the language model is shown in Figure \ref{fig: overall}(c). Specifically, we input the textual representation $\hat{x}_0$ into the language model to generate a caption $\hat{c}$. The process of a single Transformer encoder layer can be expressed as:

\begin{align}
    h'_1 &= \mathrm{DP}(\mathrm{MHA}(\mathrm{LN}(\hat{x}_0))), \\
    h_1 &= \mathrm{DP}(\mathrm{FFN}(\mathrm{LN}(h'_1))) + h'_1, \\
    h_l^{'} &= \mathrm{DP}(\mathrm{MHA}(\mathrm{LN}(h_{l-1}))) + h_{l-1}, \label{eq:7} \\
    h_l &= \mathrm{DP}(\mathrm{FFN}(\mathrm{LN}(h_l^{'}))) + h_l^{'}, \label{eq:8}
\end{align}
where $h'_1$ denotes the intermediate textual representation in first Transformer layer, $h_1$ denotes the output textual representation of first Transformer layer, $h'_l$ denotes the intermediate textual representation in the $l^{th}$ Transformer layer, $h_l$ denotes the output textual representation of the $l^{th}$ Transformer layer, and $l \in (2, 3, ..., 6)$. $\mathrm{DP}(\cdot)$ denotes drop-path, $\mathrm{FFN}(\cdot)$ denotes the feed-forward neural network in Transformer model, and $\mathrm{MHA}(\cdot)$ denotes multi-head self attention. 

Next, we input the textual representation $h_6$ output by last Transformer Encoder layer into a linear projection and a softmax. Finally, based on the vocabulary, we get the captions $\hat{c}$. This process can be expressed as:
\begin{equation}
    s = \mathrm{Softmax}(\mathrm{FC}(h_6)),
\end{equation}

\begin{equation}
    w_i = V\Big[\underset{j}{\arg\max}~s_i[j]\Big],
\end{equation}

\begin{equation}
    \hat{c}=\left(w_1, w_2, \ldots, w_T\right),
\end{equation}
where $s$ denotes the sentence representation, $s_i$ denotes the $i^{th}$ word vector, $j$ denotes the index of word vector, $V$ denotes the vocabulary vector, $w_i$ denots the $i^{th}$ word in the caption, and $T$ denotes the length of the caption.

\subsection{Objective Function}

We use the cross-entropy loss $\mathcal{L}_{ce}$ and MSE loss $\mathcal{L}_{mse}$ to guide model training, they can be expressed as:

\begin{equation}
\mathcal{L}_{mse} =\left\|f_\phi\left(x_t, v, t\right)-x_0\right\|,
\end{equation}

\begin{equation}
\mathcal{L}_{ce} = - \prod_{i=1}^n p_\theta\left(w_i \mid s_i\right),
\end{equation}
where $\phi$ denotes the parameter of Denoiser, $\theta$ denotes the parameter of DiffVC, and $n$ denotes the length of the caption.

The final loss function can be expressed as:
\begin{equation}
\mathcal{L} =\mathcal{L}_{mse}+ \mathcal{L}_{ce}.
\end{equation}

\subsection{Inference}

For inference, we directly sample noise $x_t$ from the Gaussian distribution and input it into the Denoiser. Followed DDIM \cite{song2020denoising}, to use much smaller timesteps in the inference than in the training while maintaining the quality of the generation, the inference can be expressed as:

\begin{equation}
\begin{aligned}
x_{t-1}= & \sqrt{\bar{\alpha}_{t-1}} \hat{x}_0+\sqrt{1-\bar{\alpha}_{t-1}} \epsilon_{t-1}, \\
\end{aligned}
\end{equation}
where $\sigma_t^2=\eta \tilde{\beta}_t=\eta \frac{1-\bar{\alpha}_{t-1}}{1-\bar{\alpha}_t} \beta_t$, $\eta$ is used to control the stochasticity of the sampling, and when set to $0$, the sampling possesses determinism. The probability density can be expressed as:

\begin{equation}
\begin{aligned}
& q_\sigma\left(x_{t-1} \mid x_t, x_0\right)=\mathcal{N}\left(x_{t-1} ; \sqrt{\bar{\alpha}_{t-1}} x_0+\right. \\
& \left.\sqrt{1-\bar{\alpha}_{t-1}-\sigma_t^2}\left(\frac{x_t-\sqrt{\bar{\alpha}_t} x_0}{\sqrt{1-\bar{\alpha}_t}}\right), \sigma_t^2 \mathbf{I}\right).
\end{aligned}
\end{equation} 

Thus, we can use only a subset $\tau_1, \tau_2, ..., \tau_n (n << T)$ of the training timesteps in the inference process to generate caption of the same quality, and the probability density can be expressed as:

\begin{equation}
\begin{aligned}
& q_{\sigma, \tau}\left(x_{\tau_{i-1}} \mid x_{\tau_t}, x_0\right)=\mathcal{N}\left(\mathbf{h}_{\tau_{i-1}} ; \sqrt{\bar{\alpha}_{t-1}} x_0\right. \\
& \left.+\sqrt{1-\bar{\alpha}_{t-1}-\sigma_t^2} \frac{x_{\tau_i}-\sqrt{\bar{\alpha}_t} x_0}{\sqrt{1-\bar{\alpha}_t}}, \sigma_t^2 \mathbf{I}\right).
\end{aligned}
\end{equation}

After obtaining $\hat{x}_0$ from Denoiser, the subsequent inference process is consistent with the process described in Section Non-autoregressive Language Model.

\section{Experiments}

\begin{table*}[t]
    \centering
    \caption{Video captioning model performance on MSR-VTT and MSVD datasets. - denotes that the data is not given in the corresponding literature. The bold number denotes the best results among all non-autoregressive methods.}
    \resizebox{\textwidth}{!}{
    \begin{tabular}{lc|cccc|cccc|cccc}
    \toprule
        \multirow{2}{*}{Method} & \multirow{2}{*}{Venue} & \multicolumn{4}{c|}{MSR-VTT} & \multicolumn{4}{c|}{MSVD} & \multicolumn{4}{c}{VATEX}\\
\cmidrule(lr){3-6} \cmidrule(lr){7-10} \cmidrule(lr){11-14}
& & B@4 & M & R & C & B@4 & M & R & C & B@4 & M & R & C\\
    \midrule
    \rowcolor{grey}\multicolumn{14}{l}{\textit{{Autoregressive methods}}}\\
Two-stream \cite{gao2019hierarchical} & TPAMI'20 & 39.7 & 27.0 & - & 42.1 & 54.3 & 33.5 & - & 72.8 & - & - & - & -\\
STAT \cite{yan2019stat} & TMM'20 & 39.3 & 27.1 & - & 43.8 & 52.0 & 33.3 & - & 73.8 & - & - & - & -\\
VideoTRM \cite{chen2020videotrm} & ACM MM'20 & 38.8 & 27.0 & - & 44.7 & - & - & - & - & - & - & - & -\\
STGCN \cite{pan2020spatio} & CVPR'20 & 40.5 & 28.3 & 60.9 & 47.1 & 52.2 & 36.9 & 73.9 & 93.0 & - & - & - & -\\
SAAT \cite{zheng2020syntax} & CVPR'20 & 40.5 & 28.2 & 60.9 & 49.1 & 46.5 & 33.5 & 69.4 & 81.0 & - & - & - & -\\
PMI-CAP \cite{chen2020learning} & ECCV'20 & 42.1 & 28.7 & - & 49.4 & 54.6 & 36.4 & - & 95.1 & - & - & - & -\\
ORG-TRL \cite{zhang2020object} & CVPR'20 & 43.6 & 28.8 & 62.1 & 50.9 & 54.3 & 36.4 & 73.9 & 95.2 & 32.1 & 22.2 & 48.9 & 49.7\\
SBAT \cite{jin2020sbat} & IJCAI'20 & 42.9 & 28.9 & 61.5 & 51.6 & 53.1 & 35.3 & 72.3 & 89.5 & - & - & - & -\\
TTA \cite{tu2021enhancing} & PR'21 & 41.4 & 27.7 & 61.1 & 46.7 & 52.0 & 34.0 & 70.5 & 81.2 & - & - & - & -\\
SibNet \cite{liu2018sibnet} & TPAMI'21 & 41.2 & 27.8 & 60.2 & 48.6 & 55.7 & 35.5 & 72.6 & 88.8 & - & - & - & -\\
AR-B \cite{yang2021non} & AAAI'21 & 42.0 & 28.7 & - & 49.1 & 48.7 & 35.3 & - & 91.8 & - & - & - & -\\
SGN \cite{ryu2021semantic} & AAAI'21 & 40.8 & 28.3 & 60.8 & 49.5 & 52.8 & 35.5 & 72.9 & 94.3 & - & - & - & -\\
MGRMP \cite{chen2021motion} & ICCV'21 & 41.7 & 28.9 & 62.1 & 51.4 & 55.8 & 36.9 & 74.5 & 98.5 & 34.2 & 23.5 & 50.3 & 57.6\\
FrameSel \cite{li2020adaptive} & TCSVT'22 & 38.4 & 27.2 & 59.7 & 44.1 & 50.4 & 34.2 & 70.4 & 73.7 & - & - & - & -\\
SHAN \cite{deng2021syntax} & TCSVT'22 & 39.7 & 28.3 & 60.4 & 49.0 & 54.3 & 35.3 & 72.2 & 91.3 & - & - & - & -\\
LSRT \cite{li2022long} & TIP'22 & 42.6 & 28.3 & 61.0 & 49.5 & 55.6 & 37.1 & 73.5 & 98.5 & - & - & - & -\\
TVRD \cite{wu2022towards} & TCSVT'22 & 43.0 & 28.7 & 62.2 & 51.8 & 50.5 & 34.5 & 71.7 & 84.3 & - & - & - & -\\
R-ConvED \cite{chen2023retrieval} & TOMM'23 & 40.4 & 28.1 & - & 47.9 & 53.5 & 34.6 & - & 82.4 & 32.1 & 21.8 & - & 48.7\\
EFFECT \cite{dong2023semantic} & TOMM'23 & 41.4 & 28.4 & 60.5 & 48.8 & 56.9 & 36.6 & 74.2 & 98.5 & - & - & - & -\\
RSFD \cite{zhong2023refined} & AAAI'23 & 43.4 & 29.3 & 62.3 & 53.1 & 51.2 & 35.7 & 72.9 & 96.7 & - & - & - & -\\
KG-VCN \cite{yuan2025fully} & PR'25 & 45.0 & 28.7 & 62.5 & 51.9 & 64.9 & 39.7 & 77.2 & 107.1 & 33.3 & 22.9 & 49.5 & 53.3\\
    \midrule
    \rowcolor{grey}\multicolumn{14}{l}{\textit{{Non-autoregressive methods}}}\\
    NACF \cite{yang2021non}  & AAAI'21 & 37.1 & 26.5 & 61.1 & 47.3 & 54.1 & 35.2 & 73.5 & 91.0 & - & - & - & -\\
    ALSO-Net \cite{chen2024action}  & TOMM'24 & 41.9 & 28.9 & 62.0 & 51.3 & 55.7 & 35.9 & 73.0 & 89.0 & 27.8 & 20.6 & 46.8 & 40.4\\
    \midrule
    \textbf{DiffVC (Ours)} & - & \textbf{44.5} & \textbf{31.1} & \textbf{63.9} & \textbf{56.7} & 53.5 & \textbf{37.1} & 72.5 & \textbf{95.6} & \textbf{29.5} & \textbf{21.0} & \textbf{48.7} & \textbf{50.3}\\
    \bottomrule
    \end{tabular}
    }
    \label{tab: ce}
\end{table*}

\subsection{Datasets}

MSR-VTT dataset \cite{xu2016msr} includes 10,000 videos spanning 20 distinct categories, each paired with 20 captions created by 1,327 workers. For evaluation, we use publicly available splits: 6,513 videos for training, 497 for validation, and 2,990 for testing.

MSVD dataset \cite{guadarrama2013youtube2text} consists of 1,970 short YouTube clips, each with approximately 40 English captions, totaling 70,028 annotations from Amazon Mechanical Turk workers. Videos range from 10 to 25 seconds. We split the dataset into three subsets: 1,200 videos for training, 100 for validation, and 670 for testing.

VATEX~\cite{wang2019vatex} contains 34,991 videos with 10 English annotations. The standard split includes 25,910 training videos, 3000 validation videos, and 6000 test videos.

\subsection{Metrics}

To quantitatively evaluate DiffVC, we use four established metrics: BLEU (B) \cite{papineni2002bleu}, METEOR (M) \cite{banerjee2005meteor}, ROUGE-L (R) \cite{lin2004rouge}, and CIDEr (C) \cite{vedantam2015cider}. These metrics assess the quality of generated captions by comparing them to the ground-truth sentences, with higher scores indicating better sentence generation. CIDEr is particularly valued in captioning tasks for its alignment with human judgment, while BLEU@4 (B@4) focuses on gram similarity, indicative of caption fluency. We use the standard evaluation software from MS COCO \cite{lin2014microsoft} server, with a particular emphasis on B@4 and CIDEr due to their relevance in assessing fluency and specificity, respectively.

\subsection{Implementation Details}

For video feature extraction, we follow \cite{pei2019memory} to extract spatial and temporal features to encode video information. Specifically, we use ImageNet \cite{deng2009imagenet} pre-trained ResNet-101 \cite{he2016deep} to extract 2D scene features for each frame. We also utilize Kinetics \cite{kay2017kinetics}) pre-trained ResNeXt-101 with 3D convolutions \cite{hara2018can}. For sequence length, we set it to 30 for MSR-VTT and 20 for MSVD. Optimization is performed using the Adam \cite{kingma2014adam} for 80 epochs, with the initial learning rate of 1e-4. All experiments are conducted on 8 NVIDIA V100 GPUs.

\subsection{Comparisons with State-of-the-Art Methods}

\textbf{Quantitative Comparisons.} Table \ref{tab: ce} summarizes the performance of our DiffVC on the MSR-VTT, MSVD, and VATEX. We include two groups of baselines (autoregressive and non-autoregressive). Compared with autoregressive methods, DiffVC can surpass the SOTA autoregressive methods on MSR-VTT, and achieve competitive results compared with RSFD on MSVD. Compared with non-autoregressive counterparts, DiffVC achieves best performance on all metrics on MSR-VTT and VATEX. For the MSVD, it achieves the best METEOR and CIDEr. Although DiffVC lags behind KG-VCN in MSVD, we compared their performance in generation speed and long sentence generation quality, as shown in Figures \ref{fig: intro_speed} and \ref{fig: intro_b@4}. Experimental results show that, given a fixed sentence length, DiffVC generates faster than KG-VCN, especially for long sentences, its speed advantage is significant. Furthermore, DiffVC achieves higher generation quality than KG-VCN when generating long sentences.

\begin{figure}[t]
\begin{center}
   \includegraphics[width=\linewidth]{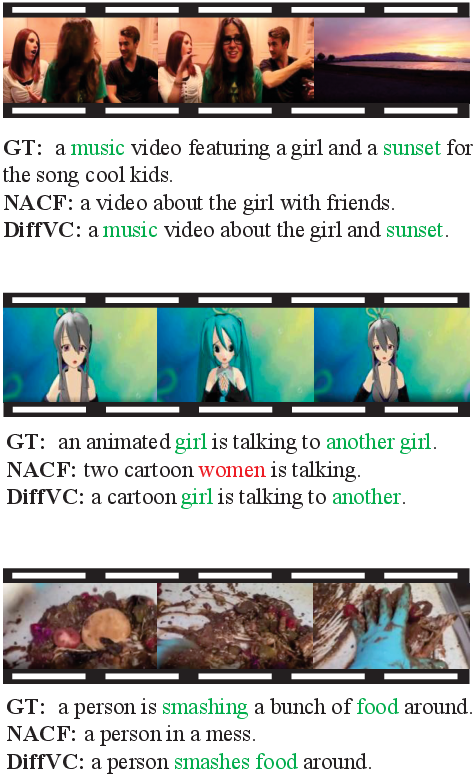}
\end{center}
   \caption{Case study on MSR-VTT for video captioning. `GT' denotes the ground-truth caption, and `NACF' denotes the caption generated by NACF\cite{yang2021non}.}
\label{fig: case}
\end{figure}

\textbf{Qualitative Comparisons.} Figure \ref{fig: case} shows the qualitative results for three samples from the MSR-VTT dataset, each described by two non-autoregressive methods (DiffVC and NACF \cite{yang2021non}) and the ground-truth captions. The content marked in red is the generated incorrect content, and the content marked in green is the generated correct content that matches the ground truth captions. The first case demonstrates DiffVC's advantage in description completeness, as it can capture low-frequency objects `sunset' in the video. The second and third cases demonstrate DiffVC's advantage in content understanding, as it can accurately distinguish between `girl' and `woman' while providing a more precise understanding of the scene content.

\subsection{Ablation Study}
\begin{table*}[t]
    \centering
    \caption{Ablation study for video captioning on MSR-VTT and MSVD datasets. The bold number denotes the best results among all methods.}
    \resizebox{0.7\textwidth}{!}{
    \begin{tabular}{lcccc|cccc}
    \toprule
        \multirow{2}{*}{Method} & \multicolumn{4}{c|}{MSR-VTT} & \multicolumn{4}{c}{MSVD} \\
\cmidrule(lr){2-5} \cmidrule(lr){6-9}
& B@4 & M & R & C & B@4 & M & R & C \\
\midrule
DiffVC & \textbf{44.5} & \textbf{31.1} & \textbf{63.9} & \textbf{56.7} & \textbf{53.5} & \textbf{37.1} & \textbf{72.5} & \textbf{95.6} \\
\textit{w/o} Discriminative Denoiser & 43.1 & 29 & 60.5 & 52.3 & 52.0 & 34.3 & 71.4 & 90.2 \\
\textit{w/o} NAR Language Model & 42.5 & 28.2 & 58.5 & 51.7 & 50.4 & 33.9 & 70.6 & 88.7\\
    \bottomrule
    \end{tabular}}
    \label{tab: ae1}
\end{table*}

We conduct ablation experiments on the MSR-VTT and MSVD datasets to evaluate the effects of various components in DiffVC.

\textbf{The Role of Discriminative Denoiser.} Table \ref{tab: ae1} quantitatively investigates the effect of the proposed Discriminative Denoiser. `\textit{w/o} Discriminative Denoiser' means the model that uses cascaded Transformer encoder blocks as the denoiser, with the rest of the model settings remaining consistent with DiffVC.

The experimental results in Table \ref{tab: ae1} show that removing the proposed discriminative denoiser significantly degrades the model's generation quality. We suppose that combining visual constraints as [class] tokens with textual tokens for self-attention calculations leads to insufficient intra-modality and inter-modality modeling, ultimately resulting in suboptimal generated text both semantically and grammatically. The proposed discriminative denoiser appropriately decouples intra-modal modeling from inter-modal modeling. Self-attention is responsible for modeling within the text modality, aiming to enhance the grammar and correctness of the text. Cross-attention is responsible for modeling the interaction between the visual and textual modalities, aiming to improve the accuracy and matching of text content.

\textbf{The Role of Non-autoregressive Language Model.} Table \ref{tab: ae1} quantitatively investigates the effect of the proposed non-autoregressive (NAR) language model. `\textit{w/o} NAR Language Model' directly inputs the textual representation output by the denoiser into a single layer of linear projection to generate the textual description. The rest of the model settings remain consistent with DiffVC.

The experimental results in Table \ref{tab: ae1} show that non-autoregressive language models significantly impact the quality of generated text. Removing the language model significantly degrades the quality of the generated text. Therefore, we suppose that it is necessary to further refine the textual representation output by the denoiser using a language model.

\begin{table}[t]
    \centering
    \caption{Ablation study on the number of denoiser blocks. The bold number denotes the best results among all methods.}
\resizebox{0.48\textwidth}{!}{
    \begin{tabular}{ccccc|cccc}
    \toprule
        \multirow{2}{*}{N} & \multicolumn{4}{c|}{MSR-VTT} & \multicolumn{4}{c}{MSVD} \\
\cmidrule(lr){2-5} \cmidrule(lr){6-9}
& B@4 & M & R & C & B-4 & M & R & C \\
\midrule
=10 & 44 & 30.5 & 61.4 & 52.1 & 53.2 & 34.3 & 70.8 & 90.2 \\
=12 & \textbf{44.5} & \textbf{31.1} & \textbf{63.9} & \textbf{56.7} & \textbf{53.5} & \textbf{37.1} & \textbf{72.5} & \textbf{95.6} \\
=14 & 44.2 & 31.1 & 63.2 & 55.8 & 52.0 & 36.1 & 72.2 & 93.7 \\
    \bottomrule
    \end{tabular}
}
    \label{tab: ae2}
\end{table}

\textbf{The Role of Denoiser Depth.} Table \ref{tab: ae2} presents an ablation study on the depth of the discriminative denoiser. We set three progressive depths of 10, 12, and 14. N denotes the number of denoiser blocks. The experimental results in Table \ref{tab: ae2} show that when the denoiser depth is set to 12, the generated text achieves the highest quality across all metrics.

\begin{table}[t]
    \centering
    \caption{Ablation study on the number of LM blocks. The bold number denotes the best results among all methods.}
\resizebox{0.48\textwidth}{!}{
    \begin{tabular}{ccccc|cccc}
    \toprule
        \multirow{2}{*}{N} & \multicolumn{4}{c|}{MSR-VTT} & \multicolumn{4}{c}{MSVD} \\
\cmidrule(lr){2-5} \cmidrule(lr){6-9}
& B@4 & M & R & C & B-4 & M & R & C \\
\midrule
=4 & 40 & 22.6 & 57.4 & 41.3 & 49.2 & 30.8 & 63.3 & 79.6 \\
=6 & \textbf{44.5} & \textbf{31.1} & \textbf{63.9} & \textbf{56.7} & \textbf{53.5} & \textbf{37.1} & \textbf{72.5} & \textbf{95.6} \\
=8 & 44.2 & 29.9 & 63.0 & 50.7 & 52.8 & 36.1 & 68.2 & 90.4 \\
=10 & 43.7 & 27.4 & 60.8 & 48.5 & 50.9 & 32.7 & 68.4 & 87.7 \\
    \bottomrule
    \end{tabular}
}
    \label{tab: ae3}
\end{table}

\textbf{The Role of the Number of LM Blocks.} To investigate the role of the number of language model blocks in a language model on model performance, we conducted ablation experiments, and the results are shown in Table~\ref{tab: ae3}. In all experimental groups, only the number of language model blocks differed. The results indicate that when the number of blocks is set to 4, the model's performance on both datasets declines significantly. This is because insufficient computation prevents the language model from generating sufficient text. When the number of blocks is set to 6, the model's performance reaches its optimal level on both datasets. Further increases in the number of blocks lead to a significant decrease in model performance. Therefore, we set the number of blocks to 6.

\begin{table}[t]
    \centering
    \caption{Ablation study on the inference step. The bold number denotes the best results among all methods.}
\resizebox{0.48\textwidth}{!}{
    \begin{tabular}{ccccc|cccc}
    \toprule
        \multirow{2}{*}{N} & \multicolumn{4}{c|}{MSR-VTT} & \multicolumn{4}{c}{MSVD} \\
\cmidrule(lr){2-5} \cmidrule(lr){6-9}
& B@4 & M & R & C & B-4 & M & R & C \\
\midrule
=5 & 41.5 & 22.6 & 57.3 & 47.7 & 51.0 & 32.4 & 67.1 & 83.5 \\
=20 & 44.5 & 31.1 & 63.9 & 56.7 & 53.5 & 37.1 & 72.5 & 95.6 \\
=50 & \textbf{45.0} & \textbf{32.3} & \textbf{64.1} & \textbf{61.8} & \textbf{53.8} & \textbf{37.2} & \textbf{73.8} & \textbf{97.}9 \\
    \bottomrule
    \end{tabular}
}
    \label{tab: ae4}
\end{table}

\textbf{The Role of the Inference Step.} To investigate the role of inference step on the performance of the Diffusion model, we conducted ablation experiments, and the results are shown in Table~\ref{tab: ae4}. In all experimental groups, only the number of language model blocks differed, and other parameters remained constant. The results indicate that the model performance generally improves gradually with increasing time steps, but the time cost also increases significantly. Therefore, we chose 20 as the time step setting to strike a balance between quality and time.

\section{Conclusion}

In this paper, we propose a diffusion-based framework for non-autoregressive video captioning, where we add Gaussian noise into the textual representation in a continuous space, derive a new textual representation from the noise according to the visual representation, and finally decode the textual representation into a textual description using a non-autoregressive language model. In addition, we propose a discriminative denoiser, which enables the model to discriminatively handle modal interactions and modeling. Extensive experiments on MSVD, MSR-VTT, and VATEX show that DiffVC solves the inherent problems in autoregression through a non-autoregressive paradigm, and the discriminative denoiser effectively improves the semantic deficiencies of non-autoregressive methods. DiffVC achieves the state-of-the-art non-autoregressive video captioning performance while being comparable to its autoregressive counterparts.

\bibliographystyle{IEEEtran}
\bibliography{references}

@String(CVPR= {IEEE Conf. Comput. Vis. Pattern Recog.})

@String(ECCV= {Eur. Conf. Comput. Vis.})

@String(AAAI = {AAAI})

@String(CVPR  = {CVPR})

@String(ECCV  = {ECCV})

@inproceedings{venugopalan2015translating,
  title={Translating Videos to Natural Language Using Deep Recurrent Neural Networks},
  author={Venugopalan, Subhashini and Xu, Huijuan and Donahue, Jeff and Rohrbach, Marcus and Mooney, Raymond and Saenko, Kate},
  booktitle={Proceedings of the 2015 Conference of the North American Chapter of the Association for Computational Linguistics: Human Language Technologies},
  pages={1494--1504},
  year={2015}
}

@inproceedings{chen2020videotrm,
  title={VideoTRM: Pre-training for video captioning challenge 2020},
  author={Chen, Jingwen and Chao, Hongyang},
  booktitle={Proceedings of the 28th ACM international conference on multimedia},
  pages={4605--4609},
  year={2020}
}

@inproceedings{liu2018sibnet,
  title={Sibnet: Sibling convolutional encoder for video captioning},
  author={Liu, Sheng and Ren, Zhou and Yuan, Junsong},
  booktitle={Proceedings of the 26th ACM international conference on Multimedia},
  pages={1425--1434},
  year={2018}
}

@inproceedings{wang2018reconstruction,
  title={Reconstruction network for video captioning},
  author={Wang, Bairui and Ma, Lin and Zhang, Wei and Liu, Wei},
  booktitle={Proceedings of the IEEE conference on computer vision and pattern recognition},
  pages={7622--7631},
  year={2018}
}

@inproceedings{wang2018m3,
  title={M3: Multimodal memory modelling for video captioning},
  author={Wang, Junbo and Wang, Wei and Huang, Yan and Wang, Liang and Tan, Tieniu},
  booktitle={Proceedings of the IEEE conference on computer vision and pattern recognition},
  pages={7512--7520},
  year={2018}
}

@inproceedings{yang2021non,
  title={Non-autoregressive coarse-to-fine video captioning},
  author={Yang, Bang and Zou, Yuexian and Liu, Fenglin and Zhang, Can},
  booktitle={Proceedings of the AAAI conference on artificial intelligence},
  volume={35},
  number={4},
  pages={3119--3127},
  year={2021}
}

@article{chen2024action,
  title={Action-aware linguistic skeleton optimization network for non-autoregressive video captioning},
  author={Chen, Shuqin and Zhong, Xian and Zhang, Yi and Zhu, Lei and Li, Ping and Yang, Xiaokang and Sheng, Bin},
  journal={ACM Transactions on Multimedia Computing, Communications and Applications},
  volume={20},
  number={10},
  pages={1--24},
  year={2024},
  publisher={ACM New York, NY}
}

@inproceedings{zhong2023refined,
  title={Refined semantic enhancement towards frequency diffusion for video captioning},
  author={Zhong, Xian and Li, Zipeng and Chen, Shuqin and Jiang, Kui and Chen, Chen and Ye, Mang},
  booktitle={Proceedings of the AAAI conference on artificial intelligence},
  volume={37},
  number={3},
  pages={3724--3732},
  year={2023}
}

@inproceedings{aafaq2019spatio,
  title={Spatio-temporal dynamics and semantic attribute enriched visual encoding for video captioning},
  author={Aafaq, Nayyer and Akhtar, Naveed and Liu, Wei and Gilani, Syed Zulqarnain and Mian, Ajmal},
  booktitle={Proceedings of the IEEE/CVF conference on computer vision and pattern recognition},
  pages={12487--12496},
  year={2019}
}

@inproceedings{chen2019motion,
  title={Motion guided spatial attention for video captioning},
  author={Chen, Shaoxiang and Jiang, Yu-Gang},
  booktitle={Proceedings of the AAAI conference on artificial intelligence},
  volume={33},
  number={01},
  pages={8191--8198},
  year={2019}
}

@article{chen2023retrieval,
  title={Retrieval augmented convolutional encoder-decoder networks for video captioning},
  author={Chen, Jingwen and Pan, Yingwei and Li, Yehao and Yao, Ting and Chao, Hongyang and Mei, Tao},
  journal={ACM Transactions on Multimedia Computing, Communications and Applications},
  volume={19},
  number={1s},
  pages={1--24},
  year={2023},
  publisher={ACM New York, NY}
}

@article{li2022long,
  title={Long short-term relation transformer with global gating for video captioning},
  author={Li, Liang and Gao, Xingyu and Deng, Jincan and Tu, Yunbin and Zha, Zheng-Jun and Huang, Qingming},
  journal={IEEE Transactions on Image Processing},
  volume={31},
  pages={2726--2738},
  year={2022},
  publisher={IEEE}
}

@article{deng2021syntax,
  title={Syntax-guided hierarchical attention network for video captioning},
  author={Deng, Jincan and Li, Liang and Zhang, Beichen and Wang, Shuhui and Zha, Zhengjun and Huang, Qingming},
  journal={IEEE Transactions on Circuits and Systems for Video Technology},
  volume={32},
  number={2},
  pages={880--892},
  year={2021},
  publisher={IEEE}
}

@article{goodfellow2020generative,
  title={Generative adversarial networks},
  author={Goodfellow, Ian and Pouget-Abadie, Jean and Mirza, Mehdi and Xu, Bing and Warde-Farley, David and Ozair, Sherjil and Courville, Aaron and Bengio, Yoshua},
  journal={Communications of the ACM},
  volume={63},
  number={11},
  pages={139--144},
  year={2020},
  publisher={ACM New York, NY, USA}
}

@misc{kingma2013auto,
  title={Auto-encoding variational bayes},
  author={Kingma, Diederik P and Welling, Max and others},
  year={2013},
  publisher={Banff, Canada}
}

@article{ho2020denoising,
  title={Denoising diffusion probabilistic models},
  author={Ho, Jonathan and Jain, Ajay and Abbeel, Pieter},
  journal={NeurIPS},
  volume={33},
  pages={6840--6851},
  year={2020}
}

@inproceedings{rombach2022high,
  title={High-resolution image synthesis with latent diffusion models},
  author={Rombach, Robin and Blattmann, Andreas and Lorenz, Dominik and Esser, Patrick and Ommer, Bj{\"o}rn},
  booktitle={CVPR},
  pages={10684--10695},
  year={2022}
}

@article{podell2023sdxl,
  title={Sdxl: Improving latent diffusion models for high-resolution image synthesis},
  author={Podell, Dustin and English, Zion and Lacey, Kyle and Blattmann, Andreas and Dockhorn, Tim and M{\"u}ller, Jonas and Penna, Joe and Rombach, Robin},
  journal={arXiv preprint arXiv:2307.01952},
  year={2023}
}

@inproceedings{sauer2025adversarial,
  title={Adversarial diffusion distillation},
  author={Sauer, Axel and Lorenz, Dominik and Blattmann, Andreas and Rombach, Robin},
  booktitle={ECCV},
  pages={87--103},
  year={2025},
  organization={Springer}
}

@inproceedings{xu2016msr,
  title={Msr-vtt: A large video description dataset for bridging video and language},
  author={Xu, Jun and Mei, Tao and Yao, Ting and Rui, Yong},
  booktitle={Proceedings of the IEEE conference on computer vision and pattern recognition},
  pages={5288--5296},
  year={2016}
}

@inproceedings{guadarrama2013youtube2text,
  title={Youtube2text: Recognizing and describing arbitrary activities using semantic hierarchies and zero-shot recognition},
  author={Guadarrama, Sergio and Krishnamoorthy, Niveda and Malkarnenkar, Girish and Venugopalan, Subhashini and Mooney, Raymond and Darrell, Trevor and Saenko, Kate},
  booktitle={Proceedings of the IEEE international conference on computer vision},
  pages={2712--2719},
  year={2013}
}

@inproceedings{papineni2002bleu,
  title={Bleu: a method for automatic evaluation of machine translation},
  author={Papineni, Kishore and Roukos, Salim and Ward, Todd and Zhu, Wei-Jing},
  booktitle={Proceedings of the 40th annual meeting of the Association for Computational Linguistics},
  pages={311--318},
  year={2002}
}

@inproceedings{banerjee2005meteor,
  title={METEOR: An automatic metric for MT evaluation with improved correlation with human judgments},
  author={Banerjee, Satanjeev and Lavie, Alon},
  booktitle={Proceedings of the acl workshop on intrinsic and extrinsic evaluation measures for machine translation and/or summarization},
  pages={65--72},
  year={2005}
}

@inproceedings{lin2004rouge,
  title={Rouge: A package for automatic evaluation of summaries},
  author={Lin, Chin-Yew},
  booktitle={Text summarization branches out},
  pages={74--81},
  year={2004}
}

@inproceedings{vedantam2015cider,
  title={Cider: Consensus-based image description evaluation},
  author={Vedantam, Ramakrishna and Lawrence Zitnick, C and Parikh, Devi},
  booktitle={Proceedings of the IEEE conference on computer vision and pattern recognition},
  pages={4566--4575},
  year={2015}
}

@inproceedings{lin2014microsoft,
  title={Microsoft coco: Common objects in context},
  author={Lin, Tsung-Yi and Maire, Michael and Belongie, Serge and Hays, James and Perona, Pietro and Ramanan, Deva and Doll{\'a}r, Piotr and Zitnick, C Lawrence},
  booktitle={European conference on computer vision},
  pages={740--755},
  year={2014},
  organization={Springer}
}

@inproceedings{pei2019memory,
  title={Memory-attended recurrent network for video captioning},
  author={Pei, Wenjie and Zhang, Jiyuan and Wang, Xiangrong and Ke, Lei and Shen, Xiaoyong and Tai, Yu-Wing},
  booktitle={Proceedings of the IEEE/CVF conference on computer vision and pattern recognition},
  pages={8347--8356},
  year={2019}
}

@inproceedings{deng2009imagenet,
  title={Imagenet: A large-scale hierarchical image database},
  author={Deng, Jia and Dong, Wei and Socher, Richard and Li, Li-Jia and Li, Kai and Fei-Fei, Li},
  booktitle={2009 IEEE conference on computer vision and pattern recognition},
  pages={248--255},
  year={2009},
  organization={Ieee}
}

@inproceedings{he2016deep,
  title={Deep residual learning for image recognition},
  author={He, Kaiming and Zhang, Xiangyu and Ren, Shaoqing and Sun, Jian},
  booktitle={Proceedings of the IEEE conference on computer vision and pattern recognition},
  pages={770--778},
  year={2016}
}

@article{kay2017kinetics,
  title={The kinetics human action video dataset},
  author={Kay, Will and Carreira, Joao and Simonyan, Karen and Zhang, Brian and Hillier, Chloe and Vijayanarasimhan, Sudheendra and Viola, Fabio and Green, Tim and Back, Trevor and Natsev, Paul and others},
  journal={arXiv preprint arXiv:1705.06950},
  year={2017}
}

@inproceedings{hara2018can,
  title={Can spatiotemporal 3d cnns retrace the history of 2d cnns and imagenet?},
  author={Hara, Kensho and Kataoka, Hirokatsu and Satoh, Yutaka},
  booktitle={Proceedings of the IEEE conference on Computer Vision and Pattern Recognition},
  pages={6546--6555},
  year={2018}
}

@article{kingma2014adam,
  title={Adam: A method for stochastic optimization},
  author={Kingma, Diederik P and Ba, Jimmy},
  journal={arXiv preprint arXiv:1412.6980},
  year={2014}
}

@article{gao2019hierarchical,
  title={Hierarchical LSTMs with adaptive attention for visual captioning},
  author={Gao, Lianli and Li, Xiangpeng and Song, Jingkuan and Shen, Heng Tao},
  journal={IEEE transactions on pattern analysis and machine intelligence},
  volume={42},
  number={5},
  pages={1112--1131},
  year={2019},
  publisher={IEEE}
}

@article{yan2019stat,
  title={STAT: Spatial-temporal attention mechanism for video captioning},
  author={Yan, Chenggang and Tu, Yunbin and Wang, Xingzheng and Zhang, Yongbing and Hao, Xinhong and Zhang, Yongdong and Dai, Qionghai},
  journal={IEEE transactions on multimedia},
  volume={22},
  number={1},
  pages={229--241},
  year={2019},
  publisher={IEEE}
}

@inproceedings{pan2020spatio,
  title={Spatio-temporal graph for video captioning with knowledge distillation},
  author={Pan, Boxiao and Cai, Haoye and Huang, De-An and Lee, Kuan-Hui and Gaidon, Adrien and Adeli, Ehsan and Niebles, Juan Carlos},
  booktitle={Proceedings of the IEEE/CVF conference on computer vision and pattern recognition},
  pages={10870--10879},
  year={2020}
}

@inproceedings{zheng2020syntax,
  title={Syntax-aware action targeting for video captioning},
  author={Zheng, Qi and Wang, Chaoyue and Tao, Dacheng},
  booktitle={Proceedings of the IEEE/CVF conference on computer vision and pattern recognition},
  pages={13096--13105},
  year={2020}
}

@inproceedings{chen2020learning,
  title={Learning modality interaction for temporal sentence localization and event captioning in videos},
  author={Chen, Shaoxiang and Jiang, Wenhao and Liu, Wei and Jiang, Yu-Gang},
  booktitle={European Conference on Computer Vision},
  pages={333--351},
  year={2020},
  organization={Springer}
}

@inproceedings{zhang2020object,
  title={Object relational graph with teacher-recommended learning for video captioning},
  author={Zhang, Ziqi and Shi, Yaya and Yuan, Chunfeng and Li, Bing and Wang, Peijin and Hu, Weiming and Zha, Zheng-Jun},
  booktitle={Proceedings of the IEEE/CVF conference on computer vision and pattern recognition},
  pages={13278--13288},
  year={2020}
}

@article{jin2020sbat,
  title={SBAT: Video captioning with sparse boundary-aware transformer},
  author={Jin, Tao and Huang, Siyu and Chen, Ming and Li, Yingming and Zhang, Zhongfei},
  journal={arXiv preprint arXiv:2007.11888},
  year={2020}
}

@article{tu2021enhancing,
  title={Enhancing the alignment between target words and corresponding frames for video captioning},
  author={Tu, Yunbin and Zhou, Chang and Guo, Junjun and Gao, Shengxiang and Yu, Zhengtao},
  journal={Pattern Recognition},
  volume={111},
  pages={107702},
  year={2021},
  publisher={Elsevier}
}

@inproceedings{ryu2021semantic,
  title={Semantic grouping network for video captioning},
  author={Ryu, Hobin and Kang, Sunghun and Kang, Haeyong and Yoo, Chang D},
  booktitle={proceedings of the AAAI Conference on Artificial Intelligence},
  volume={35},
  number={3},
  pages={2514--2522},
  year={2021}
}

@inproceedings{chen2021motion,
  title={Motion guided region message passing for video captioning},
  author={Chen, Shaoxiang and Jiang, Yu-Gang},
  booktitle={Proceedings of the IEEE/CVF international conference on computer vision},
  pages={1543--1552},
  year={2021}
}

@article{li2020adaptive,
  title={Adaptive spatial location with balanced loss for video captioning},
  author={Li, Linghui and Zhang, Yongdong and Tang, Sheng and Xie, Lingxi and Li, Xiaoyong and Tian, Qi},
  journal={IEEE Transactions on Circuits and Systems for Video Technology},
  volume={32},
  number={1},
  pages={17--30},
  year={2020},
  publisher={IEEE}
}

@article{wu2022towards,
  title={Towards knowledge-aware video captioning via transitive visual relationship detection},
  author={Wu, Bofeng and Niu, Guocheng and Yu, Jun and Xiao, Xinyan and Zhang, Jian and Wu, Hua},
  journal={IEEE Transactions on Circuits and Systems for Video Technology},
  volume={32},
  number={10},
  pages={6753--6765},
  year={2022},
  publisher={IEEE}
}

@article{dong2023semantic,
  title={Semantic embedding guided attention with explicit visual feature fusion for video captioning},
  author={Dong, Shanshan and Niu, Tianzi and Luo, Xin and Liu, Wu and Xu, Xinshun},
  journal={ACM Transactions on Multimedia Computing, Communications and Applications},
  volume={19},
  number={2},
  pages={1--18},
  year={2023},
  publisher={ACM New York, NY}
}

@article{yuan2025fully,
  title={Fully exploring object relation interaction and hidden state attention for video captioning},
  author={Yuan, Feiniu and Gu, Sipei and Zhang, Xiangfen and Fang, Zhijun},
  journal={Pattern Recognition},
  volume={159},
  pages={111138},
  year={2025},
  publisher={Elsevier}
}

@article{song2020denoising,
  title={Denoising diffusion implicit models},
  author={Song, Jiaming and Meng, Chenlin and Ermon, Stefano},
  journal={arXiv preprint arXiv:2010.02502},
  year={2020}
}

@inproceedings{wang2019vatex,
  title={Vatex: A large-scale, high-quality multilingual dataset for video-and-language research},
  author={Wang, Xin and Wu, Jiawei and Chen, Junkun and Li, Lei and Wang, Yuan-Fang and Wang, William Yang},
  booktitle={Proceedings of the IEEE/CVF international conference on computer vision},
  pages={4581--4591},
  year={2019}
}

\end{document}